\documentclass[letterpaper, 10 pt, conference]{ieeeconf} 

\IEEEoverridecommandlockouts        
\usepackage{graphics} 
\usepackage{epsfig} 
\usepackage{mathptmx} 
\usepackage{times} 
\usepackage{amsmath} 
\usepackage{amssymb}  

\usepackage[table,xcdraw]{xcolor}
\usepackage{booktabs} 
\usepackage{array} 
\usepackage{soul}
\usepackage{makecell}
\usepackage{cite}

\title{\LARGE \bf
Modular Soft Wearable Glove for Real-Time Gesture Recognition and \\ Dynamic 3D Shape Reconstruction}

\author{Huazhi Dong$^{1}$, Chunpeng Wang$^{1}$, Mingyuan Jiang$^{2}$, Francesco Giorgio-Serchi$^{2}$, and Yunjie Yang$^{1}$
\thanks{*This work was supported in part by the European Research Council Starting Grant under Grant no.101165927 (Project SELECT). (Corresponding author: Yunjie Yang)} 
\thanks{$^{1}$Huazhi Dong, Chunpeng Wang, and Yunjie Yang are with SMART Group, Institute for Imaging, Data and Communications, The University of Edinburgh, Edinburgh, UK.
        {\tt\small Huazhi.Dong@ed.ac.uk,
        s2613747@ed.ac.uk, y.yang@ed.ac.uk}}%
\thanks{$^{2}$Mingyuan Jiang and Francesco Giorgio-Serchi are with the Institute for Integrated Micro and Nano Systems, The University of Edinburgh, Edinburgh, UK.
        {\tt\small jmylxsq@163.com, F.Giorgio-Serchi@ed.ac.uk}}%
}

\begin{document}

\maketitle

\thispagestyle{empty}
\pagestyle{empty}

\begin{abstract}
With the increasing demand for human-computer interaction (HCI), flexible wearable gloves have emerged as a promising solution in virtual reality, medical rehabilitation, and industrial automation. However, the current technology still has problems like insufficient sensitivity and limited durability, which hinder its wide application. This paper presents a highly sensitive, modular, and flexible capacitive sensor based on line-shaped electrodes and liquid metal (EGaIn), integrated into a sensor module tailored to the human hand’s anatomy. The proposed system independently captures bending information from each finger joint, while additional measurements between adjacent fingers enable the recording of subtle variations in inter-finger spacing. This design enables accurate gesture recognition and dynamic hand morphological reconstruction of complex movements using point clouds. Experimental results demonstrate that our classifier based on Convolution Neural Network (CNN) and Multilayer Perceptron (MLP) achieves an accuracy of 99.15\% across 30 gestures. Meanwhile, a transformer-based Deep Neural Network (DNN) accurately reconstructs dynamic hand shapes with an Average Distance (AD) of 2.076$\pm$3.231 mm, with the reconstruction accuracy at individual key points surpassing SOTA benchmarks by 9.7\% to 64.9\%. The proposed glove shows excellent accuracy, robustness and scalability in gesture recognition and hand reconstruction, making it a promising solution for next-generation HCI systems.
\end{abstract}
\begin{figure}[t]
\centerline{\includegraphics[width=\columnwidth]{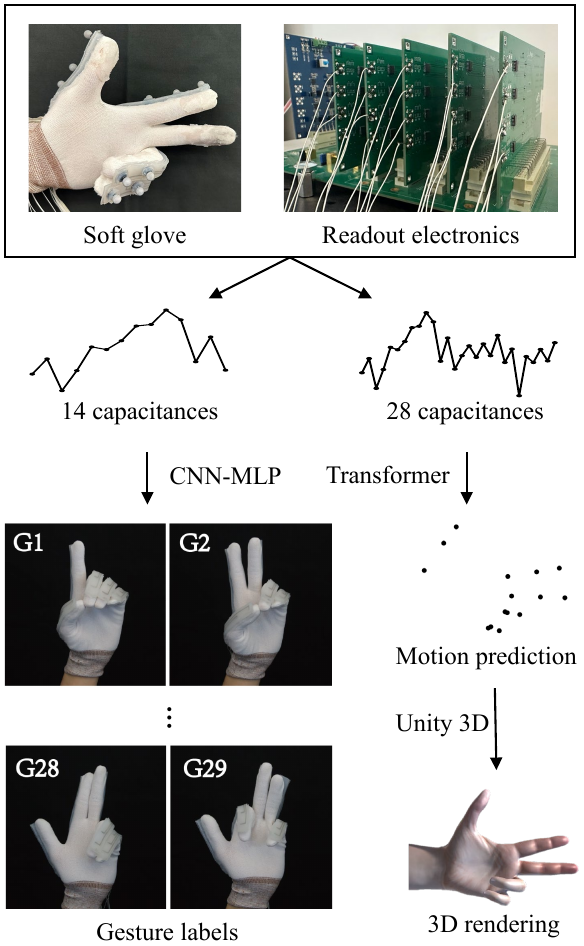}}
\caption{Schematic illustration of the proposed glove.}
\label{fig-KeyContribution}
\end{figure}

\section{INTRODUCTION}
Human hands play a pivotal role in daily tasks and professional operations due to their high degrees of freedom (DOFs) that enable exceptional flexibility and precise control \cite{gonzalez2014analysis, xiong2024single}. However, conventional rigid-body sensing technologies have inherent limitations in adaptability and responsiveness \cite{Hu2023}, making it difficult to capture the full range of hand configurations. Wearable gloves embedded with soft sensors have emerged as an effective solution for measuring the physical deformations and movements of the hand and fingers \cite{ji2024flexible}. 

Recently, several strategies have been introduced to enhance the sensing capabilities of soft gloves. One approach leverages vision-based systems using cameras and markers to capture hand poses. While this method benefits from advanced computer vision algorithms and can achieve high accuracy, its effectiveness is limited by environmental factors and occlusions \cite{ding2024vision,lee2021recon}. Another strategy utilizes the intrinsic properties of human hands, such as surface electromyography (sEMG) \cite{zhong2024sEMG,jiang2020sEmg} and crease amplification \cite{tang2021multilayered} to infer hand gestures. These approaches offer robust sensing by leveraging physiological signals, but they usually require tight skin contact, and the model needs to be trained by patient-by-patient. Consequently, variations in how users wear the glove can significantly affect the sensor's performance and effectiveness. A more direct approach employs strain sensors, which convert mechanical deformations into electrical signals. These sensors are capable of capturing hand movements without relying on the assistance of cameras or requiring a perfect fit with the hand. Such strain sensors are further categorized according to the sensing principle, including resistive\cite{Tashakori2024recon,park2024recon,Zhou2022ges,sengputa2021ges,wang2020ges,Jo2022ges,chen2021ges,Li2024ges,zhang2024recon}, capacitive \cite{wang2022ges,Atalay2017ges,glauser2019interactive,FERNANDEZ2024ges,pan2021ges,behnke2023recon}, optical \cite{li2020ultra, liu2024light} and triboelectric \cite{zhou2020ges,jiang2024ges} sensors.

\begin{figure}[t]
\centerline{\includegraphics[width=\columnwidth]
{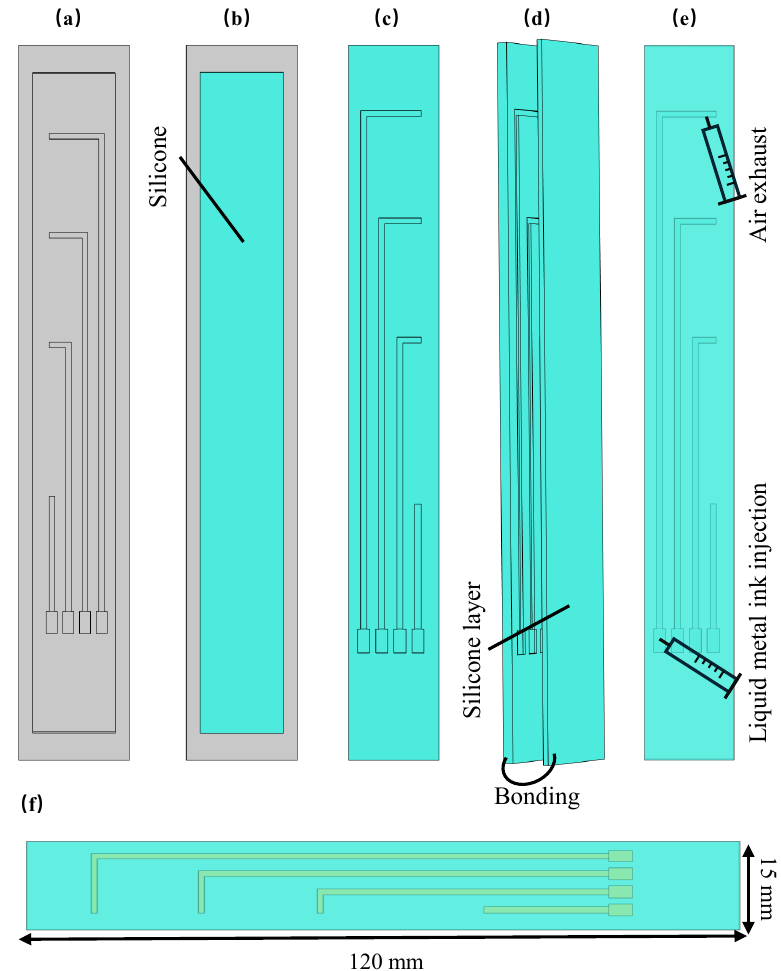}}
\caption{Fabrication of the capacitive sensor module. (a) 3D-printed mold. (b) Eco-flex 00-30 poured into the mold and cured at room temperature for three hours. (c) Substrate layer released from the mold. (d) Bonding of the substrate and sealing layers using uncured silicone mixture as adhesive. (e) Injection of liquid metal into electrode channels. (f) Final fabricated sensor module.}
\label{fig-SensorFabrication}
\end{figure}
\begin{figure}[ht]
\centerline{\includegraphics[scale=0.5]{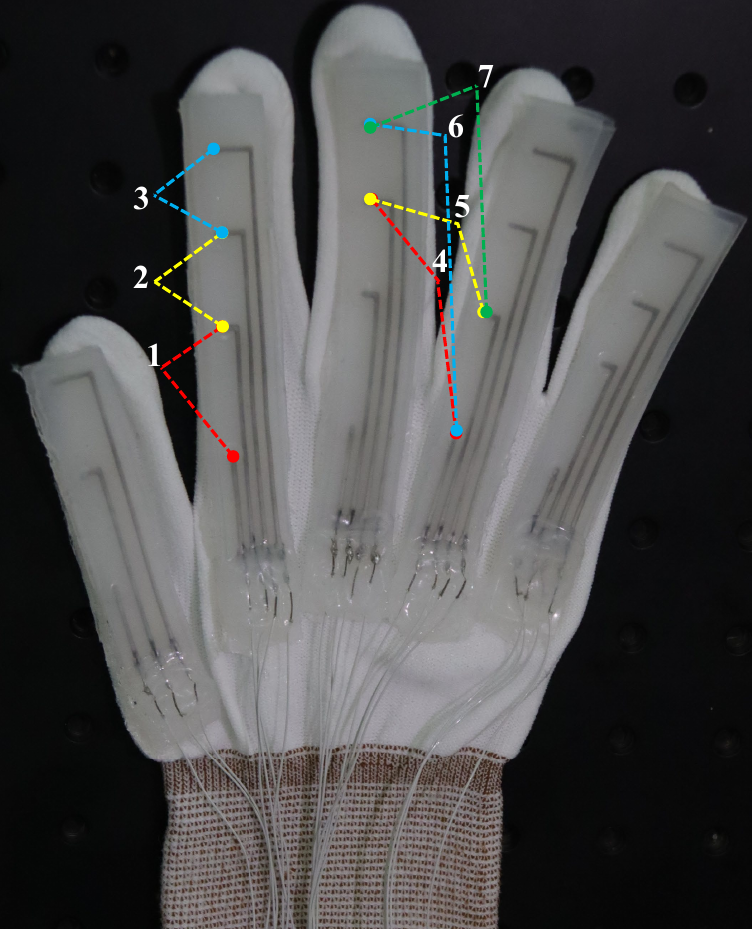}}
\caption{Assembly of the soft glove and capacitance measurement strategy.}
\label{fig-glove}
\end{figure}

Although strain sensors offer advantages over other sensing technologies, significant challenges persist in reconstructing full hand motion. Resistive sensors are easy to integrate and calibrate, yet they inherently exhibit low sensitivity. Triboelectric-based sensors can achieve self-powered sensing, but they suffer from material wear due to continuous friction during long-term use, which degrades performance and lifespan \cite{zhou2020ges}. In contrast, capacitive sensors generally provide high sensitivity, fast response, and good repeatability due to their ability to detect minute structural changes (such as variations in electrode spacing or overlap area) that alter capacitance \cite{wang2022ges, hu2024touch}. As a result, capacitive sensors have become a potential candidate for wearable sensing due to these advantages. However, the choice of materials for capacitive sensor components—including dielectric layers, electrodes, and packaging—significantly impacts performance and durability. Previous studies have explored various electrode materials for flexible capacitive sensors, including carbon black (CB) composites \cite{zhu2022carbon,zhu20203d}, carbon nanotubes (CNTs) \cite{jing2019highly}, metallic nanoparticles \cite{jung2020highly, FERNANDEZ2024ges}, and conductive textiles \cite{su2022textile}. However, these materials cannot achieve the high conductivity and stable deformation response of eutectic gallium (75.5\%)-indium (24.5\%) (EGaIn), which has a conductivity of ($3.4 \times 10^7 \mathrm{~S} \mathrm{~m}^{-1}$) and excellent mechanical adaptability \cite{Hu2023}.

Beside to sensing hardware, reconstructing the full 3D shape of the hand is challenging. Currently, most hand-shape reconstruction methods rely on sensors that measure joint angles. These sensors are typically placed only at the finger joints, capturing local curvature while neglecting the complex spatial interdependencies between adjacent fingers \cite{behnke2023recon,zhang2024recon}. Moreover, these methods often depend on kinematic models and incorporate prior information or assumptions about hand anatomy and motion patterns, which can further constrain their accuracy and adaptability \cite{park2024recon}. This limited sensor layout prevents them from fully representing the intricate and dynamic nature of hand movements. To detect joint bending and measure inter-finger spacing at the same time, earlier studies have had to deploy multiple types of sensors or deploy more sensor modules on joints and inter-finger spacing \cite{Tashakori2024recon,park2017recon,glauser2019interactive}. In contrast, our approach leverages a streamlined sensor layout and a unique sensing strategy that not only captures individual finger bending but also directly records the subtle spatial relationships between fingers.

As illustrated in Fig. \ref{fig-KeyContribution}, we present a soft wearable glove fabricated from silicone and liquid metal, designed for high-precision gesture recognition and real-time reconstruction of complex hand movements with point clouds. Our key contributions are as follows:
\begin{itemize}
\item We introduce a modular, flexible capacitive sensor tailored to the human hand, using silicone and liquid metal (EGaIn). This design offers a simplified layout with line-shaped electrodes, high sensitivity, low hysteresis, and robust durability. 
\item We propose an intra- and inter-module sensing strategy of the line-shaped electrodes capable of capturing both fine gestures and reconstructing precise overall hand movements.
\item The developed glove achieves 99.15\% accuracy in recognizing 30 distinct gestures, demonstrating an optimal balance between sensor count, sensor diversity, and gesture recognition capability. Notably, this performance is achieved with only five sensor modules, fewer or on par with other gloves, yet our system covers a larger gesture set at high accuracy.
\item The glove also enables precise tracking of hand movements and postures with an Average Distance (AD) metric of 2.076$\pm$3.231 mm in dynamic tests. Compared with SOTA \cite{park2024recon}, the reconstruction accuracy at individual key points is improved by 9.7\% up to 64.9\%. By accurately capturing both finger bending and inter-finger spacing, our system provides a highly dynamic and detailed representation of hand motion, supporting complex interactive applications.
\end{itemize}

\section{Design, Fabrication and Sensing Strategy}
\subsection{Sensor Design and Fabrication}
\begin{figure*}[ht]
\centerline{\includegraphics[width=\textwidth]{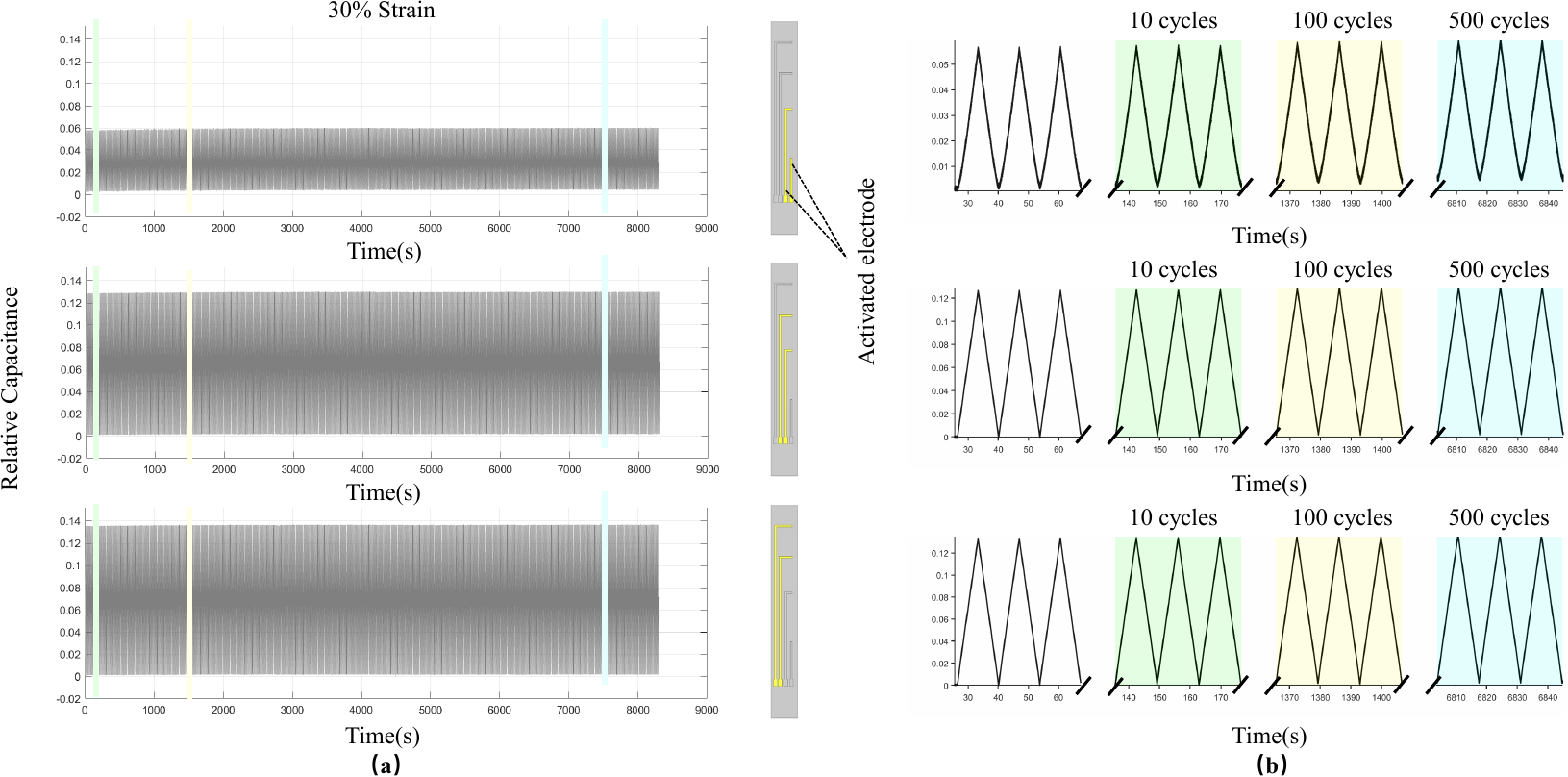}}
\caption{Cycling Test of the sensor module containing 4 line-shaped electrodes. (a) Three relative capacitance changes of the sensor module during the 3-hour cycling test. (b) The relative capacitance change after 10, 100 and 500 cycles. Relative capacitance change is defined as \(\Delta C / C_0 =  (C - C_0) / C_0\), where \(C_0\) is the initial capacitance and \(C\) is the measured capacitance under strain.}
\label{fig-cycletest}
\end{figure*}
We developed a liquid metal-based capacitive sensor module for proprioceptive sensing of finger motion. Each sensor module consists of a line-shaped electrode array composed of four parallel liquid metal traces of varying lengths embedded within a soft silicone matrix. The sensing principle is based on deformation-induced changes in capacitance. When a finger joint bends or adjacent fingers move closer together, the electrode spacing and trace length might be changed, inducing variations of the measured capacitance between the traces. The capacitance between two parallel plate electrodes is given by
\begin{equation}
C=\epsilon \cdot \frac{A}{d}
\label{eq-Q}
\end{equation}
where $C$ is the capacitance; $\epsilon$ is the permittivity of the dielectric (assumed constant); $A$ is the effective overlap area of electrodes; $d$ is the distance between electrodes.

In this design, to improve sensitivity and spatial resolution, each sensor module employs a four-electrode configuration with electrodes of graduated lengths.  The shorter electrodes are tailored to capture small-scale bending, while the longer electrodes detect global curvature changes across the finger. The electrodes are arranged in an asymmetric ‘L’ shape pattern to best capture both the chordwise and spanwise curvature of the finger (along the finger’s length and across its width). This multi-scale sensing approach ensures that even subtle variations in bending are recorded, thereby enhancing the overall performance of gesture recognition and shape reconstruction.

Fig. \ref{fig-SensorFabrication} illustrates the fabrication process and dimensions of the capacitive sensor module. Note that although the thumb and little finger modules use the same basic design as the middle finger module, their sizes (length, spacing) and number of electrodes were adjusted to account for their unique anatomy and range of motion. Finally, five fabricated sensor modules are affixed to the Nylon Glove(CR200, Polyco Healthline) using adhesive (Sil-Poxy, Smooth-On). This assembly process firmly attaches the sensors while preserving the glove’s flexibility and wearability. The sensor layout is shown in Fig. \ref{fig-glove}.

\subsection{Intra- and Inter- Module Sensing Strategy}
We designed customized capacitive sensing strategies for two different tasks to achieve an optimal balance between gesture recognition and hand morphological reconstruction.

First, for gesture recognition, we focus only on finger curvature information. As shown in Fig. \ref{fig-glove}, considering the typical kinematic structure of each finger, we configure three measurement channels for each finger (one per joint segment), and two for the thumb due to its fewer joints and unique movement. Taking the index finger as an example (marked as 1, 2, and 3), we obtain three capacitance measurements from sensors positioned at the distal interphalangeal (DIP) joint, proximal interphalangeal (PIP) joint, and metacarpophalangeal (MCP) joint. In this way, we obtain a total of 14 capacitance measurements, which can fully capture the curvature of each finger and facilitate subsequent classification and identification.

For hand morphological reconstruction, to restore the overall spatial structure of the hand more precisely, we include additional adjacent-finger measurements in addition to the 14 basic measurements. Specifically, we set up four measurements between each pair of adjacent fingers, and two measurements between the thumb and index finger. These channels measure the small changes in spacing and relative positions between neighboring fingers. Taking the middle and ring fingers as an example (marked as 4, 5, 6, and 7 in Fig. \ref{fig-glove}), we obtain four capacitance measurements from sensors placed on adjacent fingers. Through this design, a total of 28 capacitance measurements are collected for morphological reconstruction. The expanded measurement set provides more comprehensive information for the real-time reconstruction of complex hand gestures.

\subsection{Cycling Test}
We evaluated the sensor’s durability and repeatability through a 3-hour cycling test. A sensor module was subjected to repeated stretching (0\% to 30\% strain) for 3 hours. Throughout this test, the sensor’s output remained stable and highly repeatable. The capacitance readings showed a consistent, linear response at various strain levels (Fig. \ref{fig-cycletest}). Even after hundreds of cycles, there was no noticeable drift or degradation in the signal, confirming the sensor’s robustness under long-term repetitive use. We observed that a tiny 1\% strain produces approximately a 0.45\% change in the relative capacitance output. This level of sensitivity to minute strain indicates that the sensor can capture subtle finger movements or slight tactile forces.

\section{Gesture Recognition}
\subsection{Datasets}
\begin{figure}[t]
\centerline{\includegraphics[width=\columnwidth]{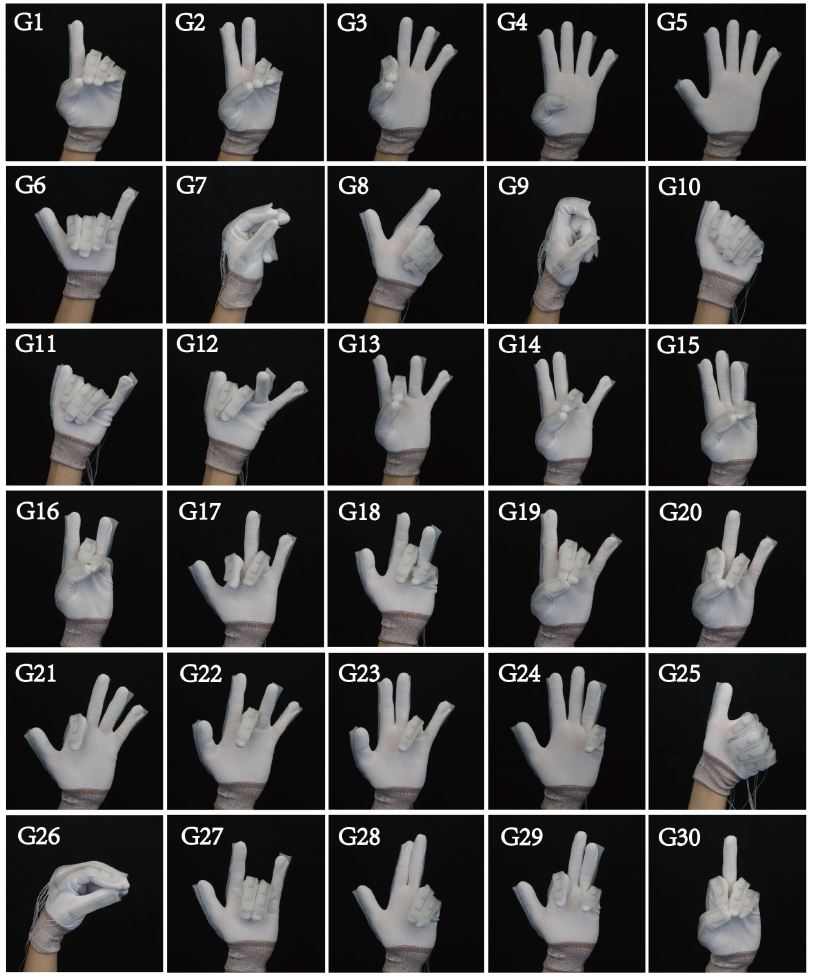}}
\caption{Pictures of 30-class hand Gestures used for gesture recognition.}
\label{fig-30Gesture}
\end{figure}
To train and evaluate the gesture recognition system, we collected a dataset from six human subjects (3 males, 3 females, 25.0 ± 5.0 years old, hand size: 16 - 19 cm). All participants provided informed consent for the experiment, data usage, and storage, with the option to withdraw at any time. As shown in Fig. \ref{fig-30Gesture}, each participant sequentially performed 30 gestures (G1-G30). Each gesture was held steady for 30 seconds without shaking or movement and then followed by 30 seconds of gentle oscillation.

Each participant contributed 210,000 frames of data (7,000 frames per gesture). Each frame contained 14 capacitive measurements and one label. The training and validation sets included four participants (2 males and 2 females), and the partition ratio was 8:2. The test set included two participants (1 male and 1 female). This dataset partitioning ensures that training and test sets evenly contain a variety of hand sizes, thereby enhancing the diversity and representativeness of the data.

\subsection{Results and Discussion}
We employ the proposed flexible sensor modules to collect the bending data of each finger, which are then processed using a classifier based on Convolution Neural Network (CNN) and Multilayer Perceptron (MLP). During training, the model constructs gesture category discrimination patterns. During testing, the system efficiently and accurately classifies input features into corresponding gesture categories, enabling real-time gesture recognition. The CNN-MLP model parameters are detailed in Table \ref{tab-trainparams}.
\begin{table}[t]
    \centering
    \caption{Model parameters}
    \setlength{\tabcolsep}{3pt} 
    \renewcommand{\arraystretch}{1.2} 
    \resizebox{\columnwidth}{!}{ 
    \begin{tabular}{lcc} 
    \toprule
    Parameter & CNN-MLP & Transformer\\ 
    \midrule
    Task & Gesture recognition & Hand reconstruction \\
    Optimizer type & Adam & Adam \\
    Base learning rate & 1.0e-4 & 1.0e-4 \\
    Total epoch & 2000 & 2000 \\
    Early stopping patience & 100 & 100\\
    Loss function & Cross Entropy Loss & Mean Squared Error\\
    Activation function & ReLU & ReLU \\
    Dropout rate  & 0.2 & 0.1\\
    Batch size  & 512 & 1024\\
    Input dimension & 14$\times$1 & 28$\times$1\\
    Output dimension & 30 & 15$\times$3\\
    Hardware & RTX 4080 GPU & RTX 4080 GPU \\
    \midrule
    Conv1d (Input, Output, Kernel Size) & [1,64,2] & - \\
    Fully connected layer 1 (Input, Output) & [832,128] & - \\
    Fully connected layer 2 (Input, Output) & [128,30] & - \\
    \midrule
    Input projection (Input, Output) & - & [28,64] \\
    Positional embedding & - & (1, 3, 64) \\
    Transformer encoder Layers & - & 3 \\
    Multi-head attention Heads & - & 2 \\
    Feedforward hidden Dimension & - & 64 \\
    Temporal window size (Frames) & - & 3 \\
    Fully connected layer (Input, Output) & - & [64, 45] \\
    \bottomrule
    \end{tabular}
    }
    \label{tab-trainparams}
\end{table}

The classification results achieved an average recognition accuracy of 99.15\%, of which the male sample reached 99.61\%, and the female sample 98.69\%, demonstrating extremely high classification consistency and robustness.   The results validate the effectiveness of the proposed sensor design and sensing strategy, which enables precise recognition of complex gestures by capturing finger curvature information from only 14 capacitance measurements. Despite minor variations across participant groups, overall classification consistency and robustness remain high, demonstrating broad applicability in diverse settings. The dynamic classification process can be observed in Supplementary Video 1. Table \ref{tab-GestureRecognition} compares our approach with state-of-the-art (SOTA) methods.  Our system achieves a balanced trade-off between sensor count, diversity, and recognition accuracy while outperforming most recent studies.
\begin{table}[ht]
    \centering
    \caption{Comparison of various studies on Gesture Recognition}
    \label{tab-GestureRecognition}
    \begin{tabular}{l|c|c|c}
    \hline
    \textbf{Study} & \makecell{\textbf{Sensor module}\\\textbf{Number}} & \makecell{\textbf{Gesture}\\\textbf{Number}}  & \textbf{Accuracy} \\
    \hline
    Pan et al. \cite{pan2021ges} & 16 & 36 & 99\% \\
    Tashakori et al. \cite{Tashakori2024recon} & 14 & 48 & 96.21\% \\
    Seyong et al. \cite{Seyong2022ges} & 5 & 8 & 99.26\% \\
    Luo et al. \cite{Luo2021} & 5 & 10 & 94\% \\
    Zhou et al. \cite{Zhou2022ges} & 5 & 11 & 98.63\% \\
    Jiang et al. \cite{jiang2024ges} & 5 & 12 & 99.2\% \\
    Faisal et al. \cite{faisal2022exploiting} & 5 & 26 & 82.19\% \\
    \hline
    \textbf{This paper} & 5 & 30 & 99.15\% \\
    \hline
    \end{tabular}
\end{table}

\section{Hand Reconstruction}
\subsection{Datasets}
\begin{figure}[t]
\centerline{\includegraphics[width=\columnwidth]{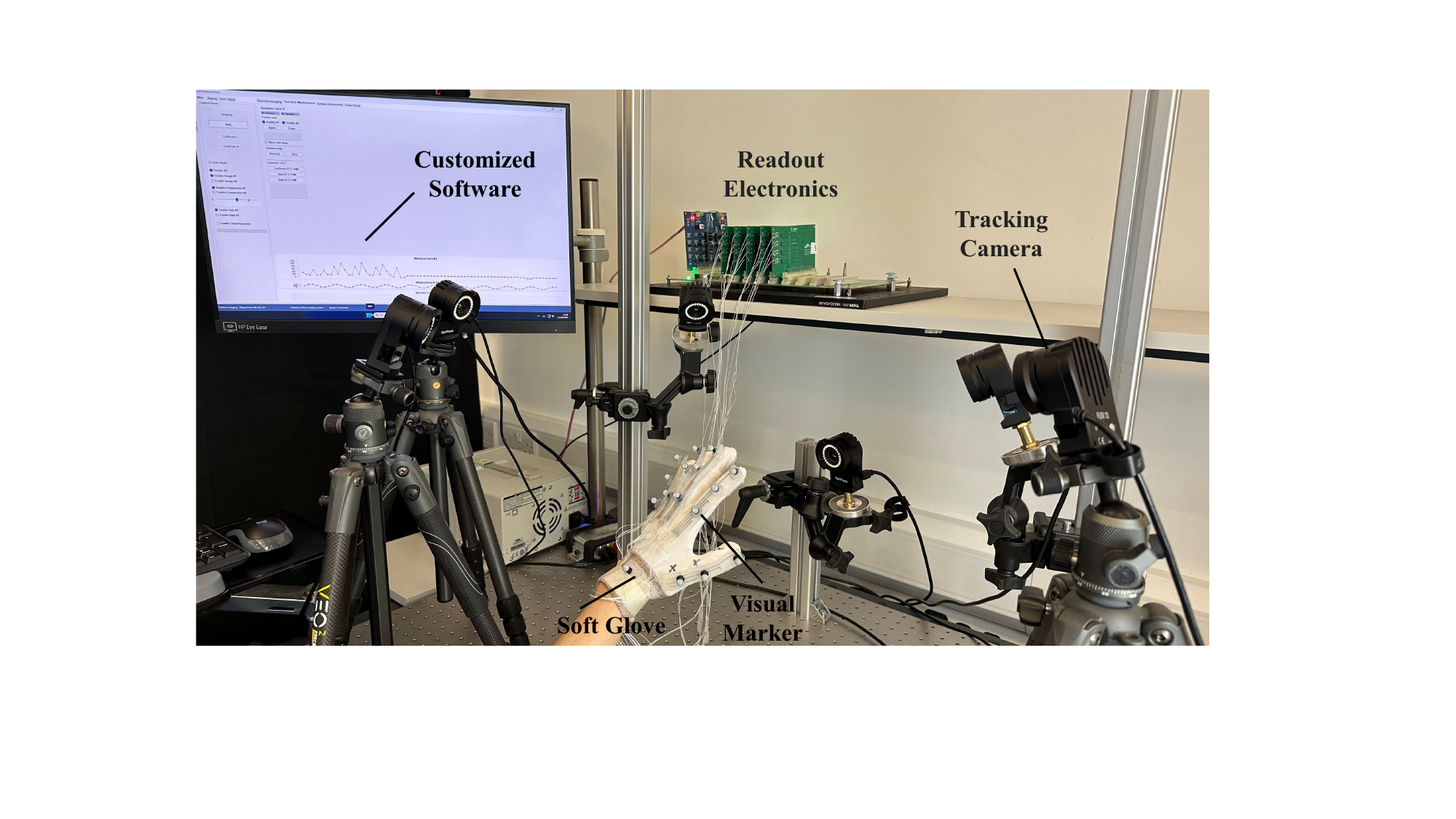}}
\caption{Experimental platform setup.}
\label{fig-TrackingCamera}
\end{figure}
Fig. \ref{fig-TrackingCamera} shows the experimental platform, which consists of the soft glove embedded with five sensor modules and 16 reflective visual markers, alongside six OptiTrack Flex 13 cameras and readout electronics. The readout electronics, developed in our previous work \cite{Hu2023}, can achieve a capacitance measurement resolution of 3 fF, and a signal-to-noise ratio exceeding 60 dB across all measurement channels. We reduced the readout electronics record data to 120 fps, consistent with the tracking camera. Notably, the marker at the wrist serves as a reference point for Unity 3D rendering and is not used as a model label.

During data collection, participants were instructed to perform random hand movements that predominantly involved two intertwined motion patterns: finger bending and adjacent finger approaching.  In natural hand movements, these two modes often occur simultaneously rather than independently.  For example, as a finger bends, its neighbouring fingers may also move closer together, leading to a coupled deformation. To capture these complex motions consistently, we recorded three separate data segments under the above motion conditions.  Each segment was used directly as one of the dataset partitions for training, validation, and testing. Finally, we collected a total of 65,324 frames of data, with each frame containing 28 capacitance measurements and a 3D point cloud with 15 points. The three segments comprised 32,965 frames (274.7 s) for training, 27,183 frames (226.5 s) for validation, and 5,446 frames (45.4 s) for testing.

\subsection{Results and Discussion}
\begin{figure*}[ht]
\centerline{\includegraphics[width=\textwidth]{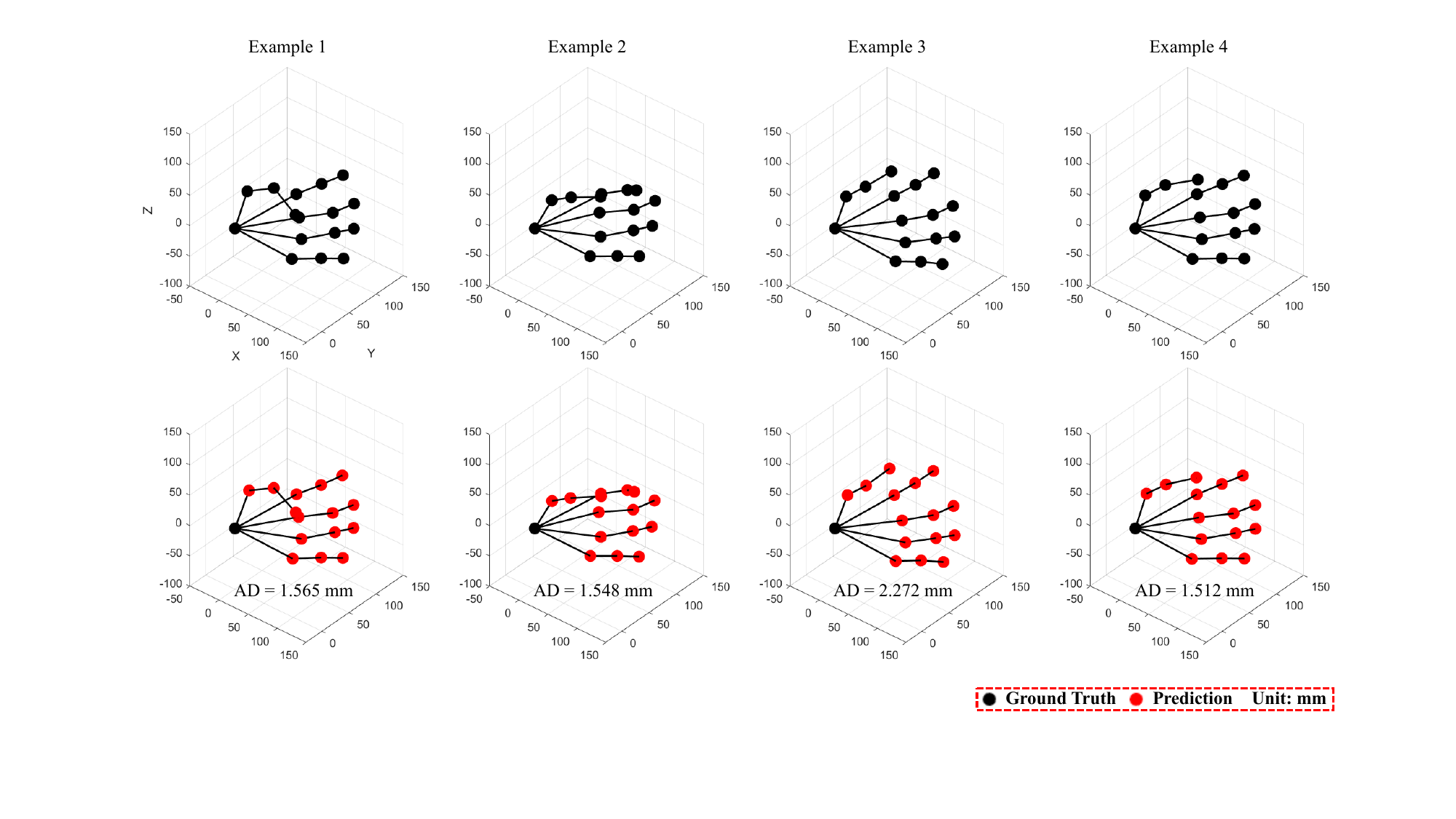}}
\caption{Hand reconstruction results. Each subfigure presents an instantaneous snapshot from a continuous hand motion sequence. The black dots indicate the ground-truth positions of reflective markers on the hand, captured by an optical tracking system. The red dots represent the corresponding positions estimated by our glove-based sensor system. The AD between the predicted and ground-truth marker positions is shown beneath each subfigure. By depicting both finger bending and inter-finger spacing, these four examples demonstrate the glove's capability to accurately track complex hand configurations in real time.}
\label{fig-handrecon}
\end{figure*}
During the data pre-processing stage, the relative capacitance and 3D point cloud data were filtered and normalized before being fed into a transformer-based DNN model. The model leverages the self-attention mechanism of the transformer to capture temporal dependencies in sensor data, enabling the prediction of long-term correlations in dynamic hand movements and facilitating accurate reconstruction of complex hand movements. Table \ref{tab-trainparams} details the model parameters.

We employ the average distance (AD) \cite{hu2024touch, dong2025learning} metric to evaluate hand movement reconstruction performance, which quantifies the discrepancy between the estimated and ground truth coordinates of visual markers:
\begin{equation}
\mathrm{AD}=\frac{1}{N M} \sum_{i=1}^N \sum_{j=1}^M \sqrt{\left(\boldsymbol{p}_{i, j}-\hat{\boldsymbol{p}}_{i, j}\right)^2}
\end{equation}
where $N$ is the number of samples in the testing set, $M$ is the number of visual markers, $\boldsymbol{p}_{i, j}$ is the ground truth coordinates of the $j^{th}$ visual marker in the $i^{th}$ testing sample and  $\hat{\boldsymbol{p}}_{i, j}$ is the estimated coordinates of the $j^{th}$ visual marker in the $i^{th}$ testing sample.

Fig. \ref{fig-handrecon} presents several examples of tracking test results, where the estimated coordinates of visual markers (red) and the ground-truth coordinates (black) are compared. Across all given cases, the estimated marker positions closely align with the ground truth, indicating high reconstruction accuracy. For the entire test dataset, the model achieves an AD of 2.076$\pm$3.231 mm. The experimental results show that the reconstructed hand shape is highly consistent with real 3D coordinates captured by the camera in key aspects such as fingertip positions, joint curvatures and inter-finger distances. The dynamic hand reconstruction process can be observed in Supplementary Video 2.

We note that the current SOTA \cite{park2024recon} is only evaluated for the 3D coordinates of a single fingertip, with an average error of 3.24 mm and 4.02 mm for two hand sizes. To better evaluate them, we calculated the average error of the 15 points in this paper from 1.412 mm to 2.9250 mm respectively, and the reconstruction accuracy of each point increased by 9.7\% to 64.9\% compared with SOTA.

\section{Conclusion}
We developed a flexible, wearable glove with liquid metal-based capacitive sensors, enabling precise capture of finger joint bending and inter-finger spacing through line-shaped electrode arrays and an inter-module sensing strategy. For gesture recognition, our system achieves 99.15\% accuracy using only 14 key measurement points. For hand morphological reconstruction, 28 composite measurements, processed by a Transformer-based DNN, enable accurate real-time hand tracking with an AD of 2.076 ± 3.231 mm, with the reconstruction accuracy at individual key points surpassing SOTA benchmarks by 9.7\% to 64.9\%. Experimental results validate the system’s robust real-time performance in both static and dynamic conditions, making it a promising solution for VR, HCI, and medical rehabilitation. Future work will focus on sensor optimization, algorithm enhancements, and deployment in more complex environments. 
\bibliographystyle{IEEEtran}
\bibliography{References}

\begin{thebibliography}{10}
\providecommand{\url}[1]{#1}
\csname url@samestyle\endcsname
\providecommand{\newblock}{\relax}
\providecommand{\bibinfo}[2]{#2}
\providecommand{\BIBentrySTDinterwordspacing}{\spaceskip=0pt\relax}
\providecommand{\BIBentryALTinterwordstretchfactor}{4}
\providecommand{\BIBentryALTinterwordspacing}{\spaceskip=\fontdimen2\font plus
\BIBentryALTinterwordstretchfactor\fontdimen3\font minus \fontdimen4\font\relax}
\providecommand{\BIBforeignlanguage}[2]{{%
\expandafter\ifx\csname l@#1\endcsname\relax
\typeout{** WARNING: IEEEtran.bst: No hyphenation pattern has been}%
\typeout{** loaded for the language `#1'. Using the pattern for}%
\typeout{** the default language instead.}%
\else
\language=\csname l@#1\endcsname
\fi
#2}}
\providecommand{\BIBdecl}{\relax}
\BIBdecl

\bibitem{gonzalez2014analysis}
F.~Gonzalez, F.~Gosselin, and W.~Bachta, ``Analysis of hand contact areas and interaction capabilities during manipulation and exploration,'' \emph{IEEE transactions on haptics}, vol.~7, no.~4, pp. 415--429, 2014.

\bibitem{xiong2024single}
Q.~Xiong, D.~Li, X.~Zhou, W.~Xin, C.~Wang, J.~W. Ambrose, and R.~Chen-Hua, ``Single-motor ultraflexible robotic (smufr) humanoid hand,'' \emph{IEEE Transactions on Medical Robotics and Bionics}, 2024.

\bibitem{Hu2023}
D.~Hu, F.~Giorgio-Serchi, S.~Zhang, and Y.~Yang, ``Stretchable e-skin and transformer enable high-resolution morphological reconstruction for soft robots,'' \emph{nature machine intelligence}, vol.~5, no.~3, pp. 261--272, 2023.

\bibitem{ji2024flexible}
B.~Ji, X.~Wang, Z.~Liang, H.~Zhang, Q.~Xia, L.~Xie, H.~Yan, F.~Sun, H.~Feng, K.~Tao \emph{et~al.}, ``Flexible strain sensor-based data glove for gesture interaction in the metaverse: a review,'' \emph{International Journal of Human--Computer Interaction}, vol.~40, no.~21, pp. 6793--6812, 2024.

\bibitem{ding2024vision}
F.~Ding, Y.~Zhu, X.~Wen, G.~Liu, and C.~X. Lu, ``Thermohands: A benchmark for 3d hand pose estimation from egocentric thermal images,'' \emph{arXiv preprint arXiv:2403.09871}, 2024.

\bibitem{lee2021recon}
Y.~Lee, W.~Do, H.~Yoon, J.~Heo, W.~Lee, and D.~Lee, ``Visual-inertial hand motion tracking with robustness against occlusion, interference, and contact,'' \emph{Science Robotics}, vol.~6, no.~58, p. eabe1315, 2021.

\bibitem{zhong2024sEMG}
X.-C. Zhong, Q.~Wang, D.~Liu, X.~Wang, R.~Li, Y.~Wang, M.~Zhang, and J.~Sun, ``Plug-and-play semg-driven hand gesture recognition with subdomain adaptation for exoskeleton rehabilitation gloves,'' \emph{IEEE Transactions on Instrumentation and Measurement}, 2024.

\bibitem{jiang2020sEmg}
S.~Jiang, L.~Li, H.~Xu, J.~Xu, G.~Gu, and P.~B. Shull, ``Stretchable e-skin patch for gesture recognition on the back of the hand,'' \emph{IEEE Transactions on Industrial Electronics}, vol.~67, no.~1, pp. 647--657, 2019.

\bibitem{tang2021multilayered}
L.~Tang, J.~Shang, and X.~Jiang, ``Multilayered electronic transfer tattoo that can enable the crease amplification effect,'' \emph{Science Advances}, vol.~7, no.~3, p. eabe3778, 2021.

\bibitem{Tashakori2024recon}
A.~Tashakori, Z.~Jiang, A.~Servati, S.~Soltanian, H.~Narayana, K.~Le, C.~Nakayama, C.-l. Yang, Z.~J. Wang, J.~J. Eng \emph{et~al.}, ``Capturing complex hand movements and object interactions using machine learning-powered stretchable smart textile gloves,'' \emph{Nature Machine Intelligence}, vol.~6, no.~1, pp. 106--118, 2024.

\bibitem{park2024recon}
M.~Park, T.~Park, S.~Park, S.~J. Yoon, S.~H. Koo, and Y.-L. Park, ``Stretchable glove for accurate and robust hand pose reconstruction based on comprehensive motion data,'' \emph{Nature communications}, vol.~15, no.~1, p. 5821, 2024.

\bibitem{Zhou2022ges}
J.~Zhou, X.~Long, J.~Huang, C.~Jiang, F.~Zhuo, C.~Guo, H.~Li, Y.~Fu, and H.~Duan, ``Multiscale and hierarchical wrinkle enhanced graphene/ecoflex sensors integrated with human-machine interfaces and cloud-platform,'' \emph{npj Flexible Electronics}, vol.~6, no.~1, p.~55, 2022.

\bibitem{sengputa2021ges}
D.~Sengupta, J.~Romano, and A.~G.~P. Kottapalli, ``Electrospun bundled carbon nanofibers for skin-inspired tactile sensing, proprioception and gesture tracking applications,'' \emph{npj Flexible Electronics}, vol.~5, no.~1, p.~29, 2021.

\bibitem{wang2020ges}
M.~Wang, Z.~Yan, T.~Wang, P.~Cai, S.~Gao, Y.~Zeng, C.~Wan, H.~Wang, L.~Pan, J.~Yu \emph{et~al.}, ``Gesture recognition using a bioinspired learning architecture that integrates visual data with somatosensory data from stretchable sensors,'' \emph{Nature Electronics}, vol.~3, no.~9, pp. 563--570, 2020.

\bibitem{Jo2022ges}
H.~S. Jo, C.-W. Park, S.~An, A.~Aldalbahi, M.~El-Newehy, S.~S. Park, A.~L. Yarin, and S.~S. Yoon, ``Wearable multifunctional soft sensor and contactless 3d scanner using supersonically sprayed silver nanowires, carbon nanotubes, zinc oxide, and pedot: Pss,'' \emph{NPG Asia Materials}, vol.~14, no.~1, p.~23, 2022.

\bibitem{chen2021ges}
X.~Chen, L.~Gong, L.~Wei, S.-C. Yeh, L.~Da~Xu, L.~Zheng, and Z.~Zou, ``A wearable hand rehabilitation system with soft gloves,'' \emph{IEEE Transactions on Industrial Informatics}, vol.~17, no.~2, pp. 943--952, 2020.

\bibitem{Li2024ges}
T.~Li, H.~Qi, Y.~Zhao, P.~Kumar, C.~Zhao, Z.~Li, X.~Dong, X.~Guo, M.~Zhao, X.~Li \emph{et~al.}, ``Robust and sensitive conductive nanocomposite hydrogel with bridge cross-linking--dominated hierarchical structural design,'' \emph{Science Advances}, vol.~10, no.~5, p. eadk6643, 2024.

\bibitem{zhang2024recon}
J.~Zhang, X.~Li, H.~Li, H.~Wang, J.~Zhang, and Y.~Hao, ``Master-slave control of rehabilitative soft glove based on collaborative sensing and fine motion recognition,'' \emph{IEEE Sensors Journal}, 2024.

\bibitem{wang2022ges}
X.~Wang, Y.~Deng, P.~Jiang, X.~Chen, and H.~Yu, ``Low-hysteresis, pressure-insensitive, and transparent capacitive strain sensor for human activity monitoring,'' \emph{Microsystems \& Nanoengineering}, vol.~8, no.~1, p. 113, 2022.

\bibitem{Atalay2017ges}
A.~Atalay, V.~Sanchez, O.~Atalay, D.~M. Vogt, F.~Haufe, R.~J. Wood, and C.~J. Walsh, ``Batch fabrication of customizable silicone-textile composite capacitive strain sensors for human motion tracking,'' \emph{Advanced Materials Technologies}, vol.~2, no.~9, p. 1700136, 2017.

\bibitem{glauser2019interactive}
O.~Glauser, S.~Wu, D.~Panozzo, O.~Hilliges, and O.~Sorkine-Hornung, ``Interactive hand pose estimation using a stretch-sensing soft glove,'' \emph{ACM Transactions on Graphics (ToG)}, vol.~38, no.~4, pp. 1--15, 2019.

\bibitem{FERNANDEZ2024ges}
F.~D.~M. Fernandez, M.~Kim, S.~Yoon, and J.~Kim, ``Capacitive batio3-pdms hand-gesture sensor: Insights into sensing mechanisms and signal classification with machine learning,'' \emph{Composites Science and Technology}, vol. 251, p. 110581, 2024.

\bibitem{pan2021ges}
J.~Pan, Y.~Li, Y.~Luo, X.~Zhang, X.~Wang, D.~L.~T. Wong, C.-H. Heng, C.-K. Tham, and A.~V.-Y. Thean, ``Hybrid-flexible bimodal sensing wearable glove system for complex hand gesture recognition,'' \emph{ACS sensors}, vol.~6, no.~11, pp. 4156--4166, 2021.

\bibitem{behnke2023recon}
L.~Behnke, L.~Sanchez-Botero, W.~R. Johnson, A.~Agrawala, and R.~Kramer-Bottiglio, ``Dynamic hand proprioception via a wearable glove with fabric sensors,'' in \emph{2023 IEEE/RSJ International Conference on Intelligent Robots and Systems (IROS)}.\hskip 1em plus 0.5em minus 0.4em\relax IEEE, 2023, pp. 149--154.

\bibitem{li2020ultra}
Z.~Li, L.~Cheng, and Q.~Song, ``An ultra-stretchable and highly sensitive photoelectric effect-based strain sensor: Implementation and applications,'' \emph{IEEE Sensors Journal}, vol.~21, no.~4, pp. 4365--4376, 2020.

\bibitem{liu2024light}
Z.~Liu, Z.~Li, J.~Wei, H.~Li, Y.~Zou, D.~Meng, and L.~Cheng, ``A light waveguide bending sensor with a high sensing range and high sensitivity,'' in \emph{2024 IEEE International Conference on Real-time Computing and Robotics (RCAR)}.\hskip 1em plus 0.5em minus 0.4em\relax IEEE, 2024, pp. 1--6.

\bibitem{zhou2020ges}
Z.~Zhou, K.~Chen, X.~Li, S.~Zhang, Y.~Wu, Y.~Zhou, K.~Meng, C.~Sun, Q.~He, W.~Fan \emph{et~al.}, ``Sign-to-speech translation using machine-learning-assisted stretchable sensor arrays,'' \emph{Nature Electronics}, vol.~3, no.~9, pp. 571--578, 2020.

\bibitem{jiang2024ges}
F.~Jiang, G.~Thangavel, J.~P. Lee, A.~Gupta, Y.~Zhang, J.~Yu, T.~Yokota, K.~Yamagishi, Y.~Zhang, T.~Someya \emph{et~al.}, ``Self-healable and stretchable perovskite-elastomer gas-solid triboelectric nanogenerator for gesture recognition and gripper sensing,'' \emph{Science Advances}, vol.~10, no.~41, p. eadq5778, 2024.

\bibitem{hu2024touch}
D.~Hu, H.~Dong, Z.~Liu, Z.~Chen, F.~Giorgio-Serchi, and Y.~Yang, ``Touch and deformation perception of soft manipulators with capacitive e-skins and deep learning,'' \emph{IEEE Sensors Journal}, 2024.

\bibitem{zhu2022carbon}
Y.~Zhu, X.~Chen, K.~Chu, X.~Wang, Z.~Hu, and H.~Su, ``Carbon black/pdms based flexible capacitive tactile sensor for multi-directional force sensing,'' \emph{Sensors}, vol.~22, no.~2, p. 628, 2022.

\bibitem{zhu20203d}
Z.~Zhu, H.~S. Park, and M.~C. McAlpine, ``3d printed deformable sensors,'' \emph{Science advances}, vol.~6, no.~25, p. eaba5575, 2020.

\bibitem{jing2019highly}
X.~Jing, H.~Li, H.-Y. Mi, Y.-J. Liu, P.-Y. Feng, Y.-M. Tan, and L.-S. Turng, ``Highly transparent, stretchable, and rapid self-healing polyvinyl alcohol/cellulose nanofibril hydrogel sensors for sensitive pressure sensing and human motion detection,'' \emph{Sensors and Actuators B: Chemical}, vol. 295, pp. 159--167, 2019.

\bibitem{jung2020highly}
Y.~Jung, W.~Lee, K.~Jung, B.~Park, J.~Park, J.~Ko, and H.~Cho, ``A highly sensitive and flexible capacitive pressure sensor based on a porous three-dimensional pdms/microsphere composite,'' \emph{Polymers}, vol.~12, no.~6, p. 1412, 2020.

\bibitem{su2022textile}
M.~Su, P.~Li, X.~Liu, D.~Wei, and J.~Yang, ``Textile-based flexible capacitive pressure sensors: A review,'' \emph{Nanomaterials}, vol.~12, no.~9, p. 1495, 2022.

\bibitem{park2017recon}
W.~Park, K.~Ro, S.~Kim, and J.~Bae, ``A soft sensor-based three-dimensional (3-d) finger motion measurement system,'' \emph{Sensors}, vol.~17, no.~2, p. 420, 2017.

\bibitem{Seyong2022ges}
S.~Oh, J.-I. Cho, B.~H. Lee, S.~Seo, J.-H. Lee, H.~Choo, K.~Heo, S.~Y. Lee, and J.-H. Park, ``Flexible artificial si-in-zn-o/ion gel synapse and its application to sensory-neuromorphic system for sign language translation,'' \emph{Science Advances}, vol.~7, no.~44, p. eabg9450, 2021.

\bibitem{Luo2021}
Y.~Luo, Y.~Li, P.~Sharma, W.~Shou, K.~Wu, M.~Foshey, B.~Li, T.~Palacios, A.~Torralba, and W.~Matusik, ``Learning human--environment interactions using conformal tactile textiles,'' \emph{Nature Electronics}, vol.~4, no.~3, pp. 193--201, 2021.

\bibitem{faisal2022exploiting}
M.~A.~A. Faisal, F.~F. Abir, M.~U. Ahmed, and M.~A.~R. Ahad, ``Exploiting domain transformation and deep learning for hand gesture recognition using a low-cost dataglove,'' \emph{Scientific Reports}, vol.~12, no.~1, p. 21446, 2022.

\bibitem{dong2025learning}
H.~Dong, X.~Wu, D.~Hu, Z.~Liu, F.~Giorgio-Serchi, and Y.~Yang, ``Learning-enhanced electronic skin for tactile sensing on deformable surface based on electrical impedance tomography,'' \emph{IEEE Transactions on Instrumentation and Measurement}, 2025.

\end{thebibliography}
\end{document}